# Oversampling Higher-Performing Minorities During Machine Learning Model Training Reduces Adverse Impact Slightly but Also Reduces Model Accuracy

Louis Hickman[1], Jason Kuruzovich[2], Vincent Ng[3], Kofi Arhin[2], & Danielle Wilson[3]

[1]Department of Psychology, Virginia Tech; The Wharton School, University of Pennsylvania

[2]Lally School of Management, Rensselaer Polytechnic Institute

[3]Department of Psychology, University of Houston

**Oversampling Higher-Performing Minorities During Machine Learning Model Training Reduces Adverse Impact Slightly but Also Reduces Model Accuracy**

*Abstract*. Organizations are increasingly adopting machine learning (ML) for personnel assessment. However, concerns exist about fairness in designing and implementing ML assessments. Supervised ML models are trained to model patterns in data, meaning ML models tend to yield predictions that reflect subgroup differences in applicant attributes in the training data, regardless of the underlying cause of subgroup differences. In this study, we systematically under- and oversampled minority (Black and Hispanic) applicants to manipulate adverse impact ratios in training data and investigated how training data adverse impact ratios affect ML model adverse impact and accuracy. We used self-reports and interview transcripts from job applicants ($N = 2,501$) to train 9,702 ML models to predict screening decisions. We found that training data adverse impact related linearly to ML model adverse impact. However, removing adverse impact from training data only slightly reduced ML model adverse impact and tended to negatively affect ML model accuracy. We observed consistent effects across self-reports and interview transcripts, whether oversampling real (i.e., bootstrapping) or synthetic observations. As our study relied on limited predictor sets from one organization, the observed effects on adverse impact may be attenuated among more accurate ML models.

**Oversampling Higher-Performing Minorities During Machine Learning Model Training Reduces Adverse Impact Slightly but Also Reduces Model Accuracy**

"Is it going to have a disparate impact on different protected classes? That is the number one thing employers using artificial intelligence should be looking out for."
-EEOC Commissioner, Keith E. Sonderling (Strong, 2021)

Organizations are rapidly adopting tools that use artificial intelligence and machine learning (ML) for many purposes, including personnel assessment and selection (e.g., Campion et al., 2016; Hickman et al., 2022; Langer et al., 2020). However, significant concerns have been raised throughout society regarding the fairness and ethicality of ML assessments (Landers & Behrend, 2022; Tippins et al., 2021). In the United States, a key legal concern for ML assessments is that personnel selection decisions that cause adverse (or *disparate*) impact—substantially different hiring rates between groups that disadvantage a legally protected group (Civil Rights Act, 1964)—constitute *prima facie* evidence of employment discrimination.

Several algorithmic solutions that adjust models to achieve equal group outcomes have been proposed to address group disparities in ML assessments (e.g., Calmon et al., 2017; Hardt et al., 2016; Kamishima et al., 2012; Kleinberg et al., 2018a; Zemel et al., 2013), but many provide the final ML model with demographic information explicitly (e.g., by using demography as a predictor) or implicitly (e.g., by creating separate models for each group) during test administration. Both are likely illegal in the United States because they constitute disparate treatment (Civil Rights Act, 1964) and/or subgroup norming (Civil Rights Act, 1991) during test administration. Therefore, there is a pressing need to advance our understanding of the causes of and potential (legal) remedies to ML model adverse impact.

ML models tend to reflect subgroup differences in applicant attributes in the training data, which are then reflected in the ML model predictions. We investigate whether this tendency can be used to our advantage by examining whether removing (i.e., equal selection ratios) or reversing (i.e., selection ratios flipped to favor disadvantaged group members) subgroup differences in the training data reduces ML model adverse impact without sacrificing accuracy. To do so, we utilize a data preprocessing approach known as oversampling—techniques for resampling observations to address class imbalances (Chawla et al., 2002; Yan et al., 2020)—to manipulate adverse impact ratios in the training data. Then, we systematically examine how this affects the adverse impact and accuracy of ML models that use self-reports and interview transcripts to predict historical screening decisions.

      The present study contributes to the literature on employment discrimination in several ways. First, we answer the special issue call to investigate adverse impact in artificial intelligence and ML personnel selection systems (Campion & Campion, 2020). Second, we answer calls to test the effects of oversampling minority groups to enhance diversity in training data (Hickman et al., 2022). We do so in a real-world, high-stakes dataset where adverse impact and group representation can be directly evaluated and altered. Oversampling to balance means and sample sizes has been shown to have small positive effects on ML model measurement bias (Yan et al., 2020) defined as equal accuracy across groups (Tay et al., 2022), but we are unaware of any studies of oversampling's effects on adverse impact. By doing so with both self-reports and interview transcripts, our study addresses the fairness of both traditional and modern selection systems. Further, we investigate the effects across a variety of text mining vectorization techniques and machine learning algorithms. This allows us to estimate the effect of oversampling on adverse impact across a variety of ML modeling approaches, reducing the chances that any observed effects are algorithm-bound. Third, we compare multiple oversampling strategies to inform future research and practice. Specifically, we compare the effects of (a) adjusting training data adverse impact vs. adjusting training data adverse impact

*and* equalizing sample sizes, as well as (b) oversampling real vs. synthetic applicants. Doing so provides nuanced answers regarding how different oversampling methods affect the adverse impact and accuracy of ML model screening decisions.

## Indices of Adverse Impact

Adverse impact is often operationalized as an adverse impact (AI) ratio—or the ratio of the selection ratios of two subgroups. Selection ratios (SRs) are calculated as the number of applicants hired in a subgroup divided by the total number of applicants from that subgroup. The AI ratio is calculated by dividing one subgroup's SR by another subgroup's SR.

Adverse impact is concerned with equality of outcomes. The most common standard for identifying practically significant adverse impact and *prima facie* evidence of discrimination is the four-fifths rule, or that the SR of members of one legally protected subgroup should not be less than four-fifths the SR of members of another subgroup (Equal Employment Opportunity Commission, 1978).[1] Therefore, AI ratios should exceed .80. The AI ratio indicates the effect size of group differences in SRs and is commonly used, although significance testing is also relevant to discrimination claims (Morris, 2016). We chose to focus on the AI ratio because even minor subgroup differences in SRs become statistically significant when sample size is in the thousands, as in the present study.

## Origins of Discrimination in Machine Learning Models

ML models and their predictions reflect existing patterns in their training data. Therefore, to the extent that discrimination and/or adverse impact exist in the personnel data used to train ML models, the ML models may reflect those historical patterns (Barocas & Selbst, 2016). We now turn to summarize the standard ML model development and evaluation process, as illustrated in Figure S1, and then explain the relevant sources of ML adverse impact that motivate our oversampling approach.

---

[1] Even when selection procedures violate the four-fifths rule, employers can demonstrate the job relevance and business necessity of the selection procedure (Civil Rights Act, 1964). However, employers may also want to reduce adverse impact for ethical reasons and to reduce the likelihood of litigation (Oswald et al., 2016).

**Supervised Machine Learning in Personnel Assessment**

Most ML assessments rely on supervised ML, which involves training an algorithm to predict some known individual-level outcome, such as historical screening or hiring decisions. To do so, individual behavior must be observed in an evaluative situation. Human observers then, either using the *in situ* behavior or a more holistic process involving additional information (e.g., resumés, cover letters), rate applicants and/or make selection decisions. A machine "perceiver" then observes and quantifies individual behavior, whether this behavior is performance in an evaluative situation (e.g., an interview), on a self-report scale, or on a test (e.g., of cognitive ability). For example, in automatically scored interviews, the unstructured, natural language of interviewee responses is transcribed, vectorized, and used in an algorithm to predict the outcome of interest (e.g., Hickman et al., 2022).

During ML model development, researchers often test multiple predictor-algorithm combinations. For example, in text mining, researchers may try out multiple vectorization techniques (i.e., methods for quantifying unstructured text data, such as closed and open vocabulary; Kern et al., 2016). To do so, the data are split into training and test datasets (e.g., Year 1 and Year 2, or *k*-fold cross-validation), the algorithm is fitted (or trained) on the training data, and the resulting ML model's accuracy is estimated on the test dataset. The predictor-algorithm combination with the highest cross-validated accuracy is often trained on all available data (i.e., both the training and test data). This final ML model is applied to future, unseen cases.

However, group differences in training data may affect the ML model, as reflected in the model parameters and its predictions. Two training data disparities affect ML models: (1) *group mean differences* on the outcome variable; and (2) *differential representation*, or underrepresentation of a subgroup (e.g., Barocas & Selbst, 2016; Kleinberg et al., 2018b). Regarding group mean differences,[2] the concern is that if group mean differences in the training data are not representative of the population of applicants to which the model will be applied, this may alter the model weights in a way that favors one group of applicants over another (Barocas & Selbst, 2016). In our study, group means equal their SRs, and therefore, group mean differences represent adverse impact. We expect that training data AI ratios will affect ML model AI ratios, such that ML models trained with equal subgroup SRs (AI ratios = 1) or with subgroup SRs favoring the subgroup with the lower SR in the observed data (AI ratios > 1) will likely exhibit less adverse impact than models trained on data where AI ratios < 1.

Differential representation occurs when subgroups are unevenly represented in training data. In cases where the predictor-outcome relationships differ across groups,[3] unequal representation in the data will cause the algorithm to primarily reflect the most prevalent patterns in the training data—or those of majority group members (Barocas & Selbst, 2016). Therefore, we also investigate the effects of equally representing subgroups in training data, as it has been proposed elsewhere as a way of enhancing fairness and validity (Kleinberg et al., 2018b).

**Diversity-Validity Tradeoff: Machine Learning Edition**

Making adjustments during ML model training to enhance fairness may negatively affect the model's convergent validity (in our study, accuracy; Barocas & Selbst, 2016). The so-called diversity-validity tradeoff is analogous to this concern (Ployhart & Holtz, 2008). The tradeoff occurs because some highly valid predictors of job performance (e.g., multiple choice cognitive ability tests) exhibit large subgroup differences, whereas selection procedures with smaller subgroup differences (e.g., personality traits) tend to less validly predict job performance. One common suggestion for addressing the so-called diversity-validity tradeoff is to find equally valid selection procedures with smaller subgroup differences (Ployhart & Holtz, 2008). Therefore, if oversampling to remove adverse impact in training data enhances ML model fairness without sacrificing convergence/accuracy, doing so may be preferable to training an ML model on raw historical data.

---

[2] In the present study, we do not consider differences in standard deviations/variances between groups because the standard deviation of binary variables (like screening decisions) is determined primarily by their means. Specifically, a binary variable's standard deviation = $(np(1-p))^{.5}$ where $n$ = sample size and $p$ = the observed mean. Further, in our study, group SRs are equivalent to group means.
[3] Differential prediction is rare in selection, and when it does occur, it tends to come in the form of overpredicting minority performance (Dahlke & Sackett, 2022).

With these considerations in mind, our study addresses the following research questions:

*Research Question 1a-b*: How does adjusting the training data AI ratios affect ML model AI ratios when using a) self-report scales, b) text mined interview transcripts, or c) both self-reports and interview transcripts to predict screening decisions?

*Research Question 2*: How does equally representing subgroups (i.e., equal *N*s) in training data affect ML model AI ratios?

*Research Question 3*: How does oversampling real versus synthetic observations affect ML model AI ratios?

*Research Question 4*: How does oversampling to remove adverse impact in training data affect ML model accuracy in the test data?

## Methods

Figure 1 summarizes the present study's methods, and more detail is provided below and in the online supplement.

**Sample**

Participants in our sample applied for U.S.-based positions in a female-dominated service industry. The sample consists of 2,501 applicants (71.9% female, 36.6% White, 28.3% Black or African American, 19.1% Hispanic, 6.3% two or more races, 4.1% Asian, and the remaining demographic groups each comprised < 1% of the sample).

**Machine Learning Model Predictor Variables**

*Text Mined Interview Transcripts*

Participants recorded their answers to five interview questions using an online video platform, and computer software transcribed their responses. We applied six common vectorization techniques to convert the interviews to vectors for use as predictors in the ML models, as detailed in Figure 1 and the online supplement.

*Self-Report Survey Scores*

The self-report survey included 16 proprietary multi-item, bipolar scales developed for the purposes of job selection that measure constructs including sociability, work ethic, and analytical mindset. These self-reports were scored in two ways: (1) as raw numerical scores and (2) as percentile scores based on norms derived by the survey vendor. The online supplement reports the scales' reliability and validity.

**Outcome Variable (Screening Decisions)**

We used the organization's screening decisions as the outcome variable. The decision is binary: applicants who passed proceeded to the next stage of the hiring process (screened in), and applicants who failed did not (screened out). The overall SR = .494, White applicant SR = .60, non-White applicant SR = .43, Black applicant SR = .46, and Hispanic applicant SR = .37. These SRs result in Non-White/White AI ratio = .72, Black/White AI ratio = .77, and Hispanic/White AI ratio = .62. A baseline model that always guesses "screened out" would have accuracy = .506, and this forms the 'baseline' accuracy against which ML model accuracy is judged.

**Train and Test Data Splits for Machine Learning Models**

To compare multiple predictor-algorithm pairs, we created a stratified three-fold split of the raw data to conduct *k*-fold cross-validation. Specifically, we split the data into *k* = 3 folds such that White, Black, and Hispanic applicants had consistent *N*s and SRs in each fold, thereby maintaining the original data's properties. Across the three folds, White SR = .60, Non-White SR = .43, Black SR ranged from .45 to .47, and Hispanic SR ranged from .36 to .38. The three folds ranged in size from *N* = 832 to 835. For all experiments, we trained ML models on two folds then assessed them on the third fold and repeated the process three times, using each fold only once for testing. In total, we trained and tested: 5 (the adjusted training data AI ratios) * 2 (adjusting SRs or SRs and *N*s) * 2 (oversampling real or synthetic observations) * 154 (11 algorithms * 14 sets of predictors) + 154 (the 11 algorithms * 14 sets of predictors on the raw data) = 3,234 models in each of the three folds, or 9,702 models trained and tested. The output of our experiments used in all analyses in the manuscript and online supplement is available on OSF: https://osf.io/c46sp/?view_only=2ffb2172f8274968bf720429812deae4

**Algorithms**

We trained a variety of common machine learning algorithms, as detailed in Figure 1 and the online supplement. No one algorithm is optimal for all tasks (i.e., no free lunch theorem; Wolpert & Macready, 2005), and these represent a sample of commonly used ML algorithms. We conducted hyperparameter tuning for each algorithm in each set of training folds of the original data, as detailed in the online supplement. We fully crossed these algorithms with the two methods for scoring the self-reports, the six text mining approaches applied to the interview transcripts, and the combined predictor set of the six text mining approaches plus the raw self-report scores. This allowed us to estimate the effect of oversampling on model outcomes across a variety of predictor-algorithm pairs, thereby ensuring that our results generalize across many ML models. Tables S8-S10 report the average accuracy obtained on the raw data when SR = .50 for all algorithms, predictors, and predictor-algorithm pairs, respectively.

## Oversampling Ratios

We used under- and oversampling to investigate the relationship between training data AI ratios and ML model AI ratios in the test data. Prior to model training, we under- and oversampled minorities in the training data to achieve Black/White and Hispanic/White AI ratios ranging from .60 to 1.40, stepping by .20. In all cases, we kept Black and Hispanic SRs in the numerator. To achieve training data AI ratio = .60, we *under*sampled passing Black and Hispanic applicants to reduce their SRs. To achieve AI ratios = .80 to 1.40, we oversampled passing (i.e., screened in) Black and Hispanic applicants until the desired AI ratio was achieved.

## Oversampling Strategies

We investigated two oversampling strategies: 1) oversampling to adjust SRs and 2) oversampling to adjust SRs and equalize sample sizes. The former case is described in the "Oversampling Ratios" subsection. In doing so, Black and Hispanic sample sizes increased by the number of cases added to achieve the manipulated AI ratio. To equalize sample sizes, we multiplied the White $N$ by the desired SR (as determined by the desired AI ratio),[4] then we a) oversampled passing Black and Hispanic applicants (respectively) to reach those values and b) oversampled (or undersampled, if necessary) Black and Hispanic applicants who failed until White, Black, and Hispanic applicant $N$s were equal.

## Oversampling Techniques

We: 1) oversampled real observations with replacement or 2) oversampled synthetically generated observations. We used the Synthetic Minority Over-Sampling TEchnique (SMOTE;

---

[4] The "fail" Black applicants was $N = 4$ larger than the "fail" White applicants. To equalize sample sizes, we first oversampled four of the "fail" White applicants before oversampling Black and Hispanic applicants.

Chawla et al., 2002) to generate synthetic a) screened in Black applicants, b) screened in Hispanic applicants, c) screened out Black applicants, and d) screened out Hispanic applicants. Figure S2 illustrates how SMOTE works. We used the first two categories to adjust AI ratios and all four categories when adjusting AI ratios *and* equalizing sample sizes.

**Test Data Selection Ratio**

As a robustness check, we analyzed our results at different overall SRs in the test data: .10 and .50. To do so, we had the ML models output class probabilities (i.e., continuous values ranging from 0-1) instead of binary predictions. Then, to achieve overall SRs = .10 and .50, we set the highest 10% and 50%, respectively, of the class probabilities to 1 (pass/screen in) and the remaining values to 0 (fail/screen out). AI ratios are more likely to violate the four-fifths rule as the overall SR decreases (Oswald et al., 2016). SR = .50 is very similar to the observed SR in our data, and SR = .10 represents a more competitive (e.g., later stage) selection procedure.

## Results

To investigate our research questions, we treated the 9,702 sets of algorithmic predictions as observations for analysis and measured their accuracy and AI ratios at overall SR = .10 and .50. In the raw data, the models that used interview transcripts to predict screening decisions tended to be more accurate than models that used self-reports as predictors, and the models that used both interview transcripts *and* self-reports tended to be no more accurate than the interview models. Among models that used: interview transcripts as predictors, $Accuracy_{Max} = 69.6\%$ (averaged across the three folds); self-reports as predictors, $Accuracy_{Max} = 60.2\%$; and combined predictor sets, $Accuracy_{Max} = 70.6\%$.

Research Question 1 concerns the effect of training data AI ratios on ML model AI ratios when screening applicants in the test data. Table 1 reports the average ML model accuracy and AI ratios in the raw data and at each manipulated training data AI ratio for models that used self-reports (top), interview transcripts (middle), and both interview transcripts and self-reports (bottom) as predictors (Tables S5-S7 report the same information at overall SR = .10). On average, among models that used self-reports as predictors, changing training data AI ratios from .6 to 1.4 caused the ML model AI ratios to increase from a minimum of .11 (Hispanic/White AI Ratios) to a maximum of .16 (Black/White AI Ratios). Among models that used interview transcripts as predictors, the average increases were smaller, ranging from a minimum of .04 (Hispanic/White AI ratios) to a maximum of .07 (Black/White AI Ratios). On average, among models that used both sets of predictors, the AI ratios increased from a minimum of .06 (Black/White AI ratios) to a maximum of .08 (Non-White/White AI ratios). These findings that the effects were largest among models that used self-reports and smallest among models that used interview transcripts as predictors align with the magnitude of correlations between training data AI ratios and ML model AI ratios reported in Table S1 for each predictor set. Thus, for all three predictor sets, training data AI ratios affect ML model AI ratios.

Notably, however, the effects were sometimes small in magnitude. Table 2 and Figure 2 report the ML model AI ratios for the most accurate model from each predictor set because these are the models likely to have been selected for subsequent use. For example, the most accurate model included the Latent Semantic Indexing (LSI) operationalization of interview transcripts plus self-reports as predictors and used linear discriminant analysis for prediction. When it was trained on the raw data where Black/White AI ratio = .77, and Hispanic/White AI ratio = .62, it exhibited an average Black/White AI ratio = .715 and Hispanic/White AI ratio = .611, whereas when it was trained on data where AI ratios were adjusted via oversampling to equal 1, it exhibited an average Black/White AI ratio = .747 and Hispanic/White AI ratio = .649.

Research Question 2 regards whether equalizing subgroup *N*s further enhances ML model AI ratios. Tables S2-S4 report the average ML model accuracy and AI ratios, respectively, for ML models that used self-reports, interview transcripts, and the combined predictor set in each experimental condition at overall SR = .50 (Tables S5-S7 report the same at overall SR = .10). AI ratios tended to increase by about .01 (although did not always do so) from manipulating SRs *and* equalizing *N*s when compared to only manipulating SRs. These findings align with the correlations between a dummy variable for whether SRs or SRs and *N*s were manipulated and the ML model AI ratios, in that the correlations show a minimal positive effect of equalizing *N*s beyond manipulating SRs. Overall, equalizing sample sizes tended to exert minimal, positive effects beyond manipulating training data AI ratios.

Research Question 3 regards whether different effects are observed from oversampling real versus synthetic observations generated by SMOTE. As reported in Tables S2-S4, these effects tended to be even smaller in magnitude than the effect of manipulating SRs versus manipulating SRs and equalizing *N*s. Further, as reported in Table S1, the effects were mixed across predictor sets, such that synthetic observations increased Non-White/White and Black/White AI ratios among models that used self-reports as predictors but decreased them and Hispanic/White AI ratios among models that used interview transcripts or both interview transcripts and self-reports as predictors. Therefore, oversampling real or synthetic observations provided similar effects.

Research Question 4 addresses the tradeoff between ML model accuracy and adverse impact. Table 2 reports the most accurate models' accuracy when trained on the raw data vs. when training data AI ratios = 1, and Figure 2 illustrates the tradeoff between accuracy and Non-White/White AI ratios for these models when training data AI ratios = .6, .8, 1.0, 1.2, and 1.4 (on average across the other conditions). As Figure 2 shows, oversampling to adjust training data AI ratios tended to slightly decrease model accuracy. For example, for the models that used LSI and self-reports as predictors, AI ratios increased by .100 when training models on training data AI ratios = 1.4 compared to training on the raw data, and accuracy decreased by .027. Among models that used only LSI as predictors, AI ratios increased by .091 when training models on training data AI ratios = 1.4 compared to training on the raw data, and accuracy decreased by .006. This aligns with trends reported in Table 1 when all models' outputs were examined.

## Discussion

Adverse impact is a foundational concern for ML-powered selection tools, as they receive heightened scrutiny from applicants and policymakers. The present study investigated the effects of oversampling high-performing minorities, a technique being explored by data and computer scientists (Yan et al., 2020), on ML model AI ratios. Removing or reversing adverse impact in training data increased ML model AI ratios while reducing ML model accuracy, although the effect sizes were small.

### Theoretical and Practical Implications

Although adequate representation in training data is important for developing ML models that are equally accurate across demographic groups (Barocas & Selbst, 2016; Kleinberg et al., 2018b), equal representation had very minor, positive effects on ML model AI ratios in our study. This may be because oversampling minority success already affected differential representation in our training data. Indeed, to create training data AI ratios = 1.0, 1.2, and 1.4, we necessarily oversampled many minority applicants, thereby increasing their sample size beyond the number of White applicants. Therefore, equal representation may have independent, positive effects, but our adjustments to training data AI ratios may have suppressed them.

Further, the observed effects were similar regardless of whether real or synthetically generated observations were oversampled. This is encouraging, because although binary classification problems tend to benefit more from oversampling synthetic than real observations (Chawla et al., 2002), there is something uncanny about using synthetic observations for personnel assessment. We encourage future work to continue to check if this holds true in other studies, but the current findings suggest that practitioners can reduce ML model adverse impact to a similar degree regardless of whether they oversample real or synthetic observations.

Although oversampling to adjust training data AI ratios slightly increased ML model AI ratios, doing so also tended to slightly reduce ML model accuracy. This issue is analogous to the so-called diversity-validity dilemma (Ployhart & Holtz, 2008) and a known limitation of methods of enhancing algorithmic fairness (Barocas & Selbst, 2016). Such decreases in accuracy limit the potential practical value of oversampling. Importantly, however, if ML models perfectly replicate historical human decisions, they could not reduce adverse impact. Future work is needed to determine how such adjustments affect validity and predictive bias, as these are more important than replicating historical decisions and our data did not include workplace outcomes.

**Limitations and Future Research**

The generalizability of our findings to other personnel selection situations is limited by two primary properties of the dataset. First, our study focused on a subset of potentially useful predictors for personnel selection (i.e., self-reports and interview responses), yet many selection systems may have a broader array of predictors available, such as biodata and cognitive ability. Second, the sample size in our study is rather small for real-world ML applications, as they may be based, in practice, on tens of thousands of observations. The relatively low accuracy of our ML models may have enhanced the magnitude of oversampling's effects, whereas more accurate models may exhibit smaller effects from oversampling. Both a broader array of predictors and a larger sample size could potentially increase ML model accuracy and alter the effects of oversampling.

Due to the small effects of oversampling, future research should investigate additional approaches for addressing ML model adverse impact. For example, removing predictors that are predictive of group membership holds potential for reducing adverse impact (Booth et al., 2021), as does reducing the weight given to such predictors (Zhang et al., 2018). Future research is needed to determine the effects of these and other approaches in high-stakes settings.

The effects of oversampling high-performing minorities rely on an assumption that subgroup differences in predictors will be consistent in new data. If, however, subgroup differences on predictors were inconsistent between the training data and subsequent applicants, then training data AI ratios may have a weaker relationship with ML model AI ratios. This suggests another route to addressing adverse impact, regardless of whether ML models are used for assessment: enacting societal change to reduce subgroup differences in job-relevant qualifications (i.e., predictors). Mean racial subgroup differences in job-relevant qualifications begin at a young age (McDaniel et al., 2011), and without change, they may persist in society for another 90 years or more (Barrett et al., 2011). Addressing subgroup differences in job-relevant qualifications is the most direct route for reducing adverse impact across assessment methods.

ADVERSE IMPACT IN TRAINING DATA    16
## References

Barocas, S., & Selbst, A. D. (2016). Big data's disparate impact. *California Law Review, 104*(3), 671-732.

Barrett, G. V., Miguel, R. F., & Doverspike, D. (2011). The Uniform Guidelines: Better the devil you know. *Industrial and Organizational Psychology*, *4*(4), 534-536.

Booth, B. M., Hickman, L., Subburaj, S. K., Tay, L., Woo, S. E., & D'Mello, S. K. (2021). Bias and fairness in multimodal machine learning: A case study of automated video interviews. *Proceedings of the 2021 International Conference on Multimodal Interaction (ICMI '21)*. https://doi.org/10.1145/3462244.3479897

Calmon, F. P., Wei, D., Vinzamuri, B., Ramamurthy, K. N., & Varshney, K. R. (2017, December). Optimized pre-processing for discrimination prevention. In *Proceedings of the 31st International Conference on Neural Information Processing Systems* (pp. 3995-4004).

Campion, M. C., Campion, M. A., Campion, E. D., & Reider, M. H. (2016). Initial investigation into computer scoring of candidate essays for personnel selection. *Journal of Applied Psychology, 101*(7), 958-975.

Campion, M. A., & Campion, E. D. (2020). *Call for papers Personnel Psychology Special Issue Part 2: Applying Machine Learning and Artificial Intelligence to Personnel Selection*.

Chawla, N. V., Bowyer, K. W., Hall, L. O., & Kegelmeyer, W. P. (2002). SMOTE: synthetic minority over-sampling technique. *Journal of Artificial Intelligence Research*, *16*, 321-357.

Civil Rights Act of 1964. (1964). Pub. L. No. 88-352, 78 Stat. 243.

Civil Rights Act of 1991. (1991). Pub. L. No. 102-166, 105 Stat. 1071.

Dahlke, J. A., & Sackett, P. R. (2022). On the assessment of predictive bias in selection systems with multiple predictors. *Journal of Applied Psychology, 107*(11), 1995–2012.

Elazar, Y., & Goldberg, Y. (2018). Adversarial removal of demographic attributes from text data. *Proceedings of the 2018 Conference on Empirical Methods in Natural Language Processing*, 11–21. https://doi.org/10.18653/v1/D18-1002

Equal Employment Opportunity Commission, Civil Service Commission, Department of Labor, & Department of Justice. (1978). *Uniform guidelines on employee selection procedures*. http://uniformguidelines.com/uniguideprint.html

Hardt, M., Recht, B. & Singer, Y. (2016). Train faster, generalize better: Stability of stochastic gradient descent. *Proceedings of The 33rd International Conference on Machine Learning*, *48*, 1225-1234.

Hickman, L., Bosch, N., Ng, V., Saef, R., Tay, L., & Woo, S. E. (2022). Automated video interview personality assessments: Reliability, validity, and generalizability investigations. *Journal of Applied Psychology*, *107*(8), 1323–1351. https://doi.org/10.1037/apl0000695

Kamishima, T., Akaho, S., Asoh, H., & Sakuma, J. (2012, September). Fairness-aware classifier with prejudice remover regularizer. In J*oint European Conference on Machine Learning and Knowledge Discovery in Databases* (pp. 35-50).

**Tables**

**Table 1**

*Average ML Model Accuracy and Adverse Impact Ratios (Overall SR = .50)*

|  |  | Accuracy |  |  |  | AI Ratio |  |  |
| --- | --- | --- | --- | --- | --- | --- | --- | --- |
|  | Train AI Ratio | Overall | White | Black | Hispanic | NW/W | B/W | H/W |
| Self-reports | Raw | .568 | .584 | .558 | .569 | .736 | .692 | .753 |
|  | .6 | .564 | .581 | .552 | .563 | .724 | .689 | .737 |
|  | .8 | .564 | .577 | .554 | .562 | .743 | .711 | .754 |
|  | 1.0 | .560 | .570 | .551 | .560 | .768 | .745 | .771 |
|  | 1.2 | .554 | .560 | .547 | .552 | .807 | .789 | .808 |
|  | 1.4 | .544 | .545 | .541 | .540 | .854 | .844 | .849 |
| Interview | Raw | .625 | .622 | .623 | .636 | .844 | .869 | .795 |
| Transcripts | .6 | .630 | .628 | .629 | .643 | .835 | .858 | .787 |
|  | .8 | .630 | .628 | .630 | .641 | .845 | .872 | .793 |
|  | 1.0 | .627 | .625 | .628 | .636 | .853 | .884 | .795 |
|  | 1.2 | .620 | .617 | .620 | .629 | .867 | .902 | .809 |
|  | 1.4 | .611 | .608 | .612 | .620 | .891 | .928 | .831 |
| Combined | Raw | .633 | .636 | .628 | .639 | .782 | .798 | .739 |
| Predictors | .6 | .631 | .635 | .624 | .642 | .768 | .785 | .726 |
|  | .8 | .631 | .633 | .627 | .640 | .781 | .801 | .735 |
|  | 1.0 | .628 | .629 | .624 | .637 | .799 | .825 | .747 |
|  | 1.2 | .623 | .623 | .621 | .631 | .818 | .848 | .762 |
|  | 1.4 | .613 | .613 | .612 | .619 | .850 | .844 | .793 |

For self-reports, $N = 66$ models on the raw data; $N = 1,320$ models when oversampling; for both interview transcripts and combined predictors, $N = 198$ models on the raw data, $N = 3,960$ models when oversampling.

**Table 2**

*Average Accuracy and Adverse Impact Ratios for Most Accurate Models (Overall SR = .50)*

|  |  |  | Test AI Ratios |  |  |
| --- | --- | --- | --- | --- | --- |
|  |  | Overall |  |  |  |
| Model | Train AI Ratio | Accuracy | NW/W | B/W | H/W |
| Self-reports (raw) | Raw | .602 | .669 | .627 | .674 |
|  | 1.0 | .587 | .725 | .715 | .720 |
| LSI | Raw | .696 | .695 | .712 | .625 |
|  | 1.0 | .697 | .716 | .742 | .647 |
| LSI + Self-reports | Raw | .706 | .688 | .715 | .611 |
|  | 1.0 | .693 | .719 | .747 | .649 |

*Note*: Results averaged across folds, oversampling methods and techniques (on the raw data, for each, $N = 3$; when train AI ratio = 1.0, $N = 12$ for each. NW/W = Non-White/White AI Ratio; B/W = Black/White AI Ratio; H/W = Hispanic/White AI Ratio. Logistic regression provided the highest accuracy for survey scores; ridge regression for LSI; and linear discriminant analysis for LSI + Self-reports.

# Figures

**Figure 1**

*Methods Overview for the Present Study*

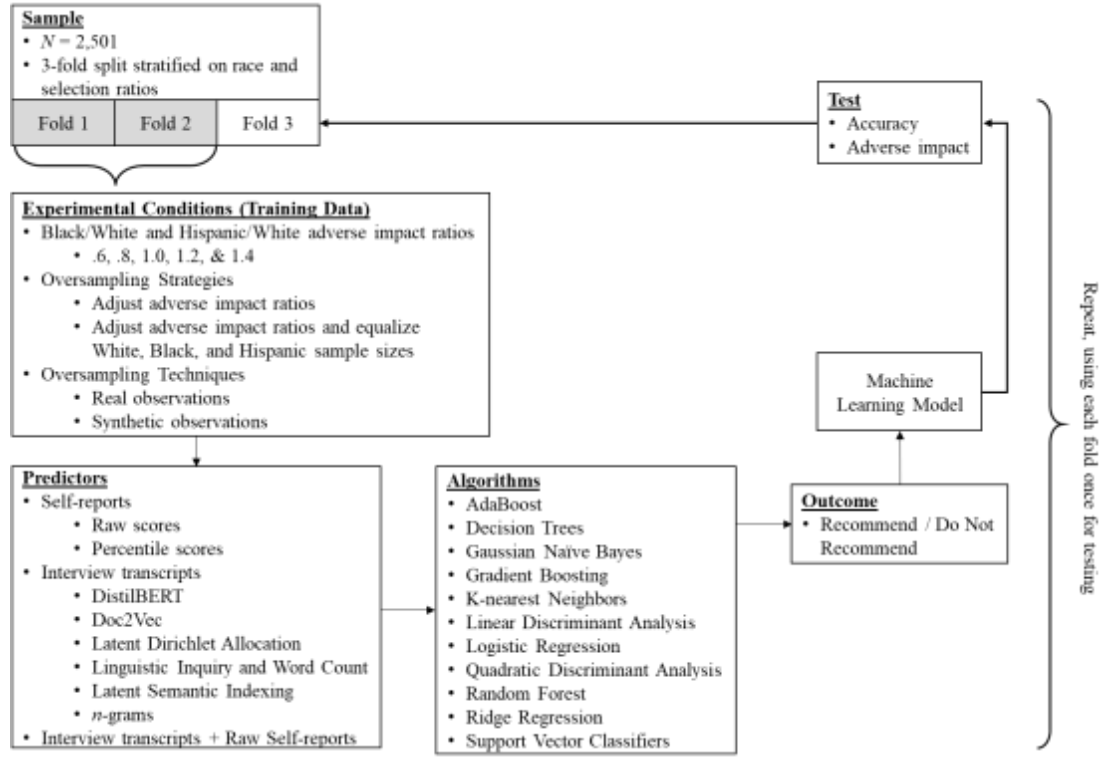

**Figure 2**

*Average Fairness-Accuracy Tradeoff for the Most Accurate Models (Selection Ratio = .50)*

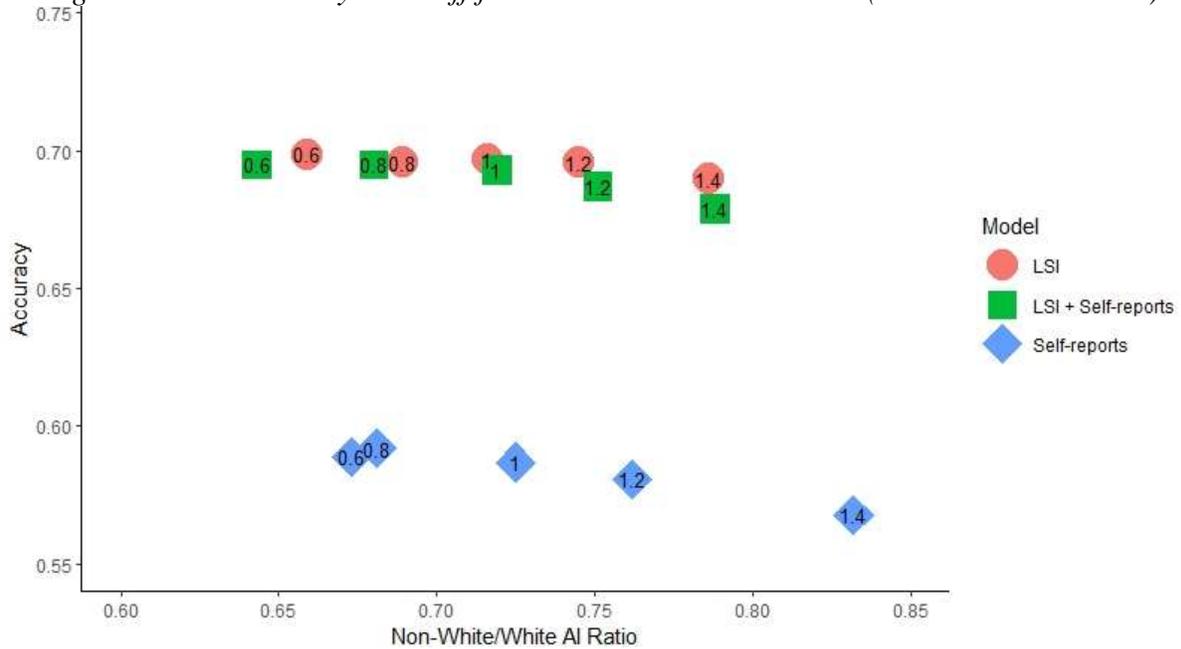

*Note*: Numbers in shapes indicate the training data Black/White and Hispanic/White AI ratios.

**Oversampling Higher-Performing Minorities During Machine Learning Model Training Reduces Adverse Impact Slightly but Also Reduces Model Accuracy**

**Online Supplement**

**Literature Review on Oversampling**

**Oversampling**

Because oversampling is not commonly used for the purposes employed in the present study in psychology and management, this section provides additional information on the technique. To adjust group selection ratios (SRs), adverse impact (AI) ratios, and sample sizes in the training data, we draw on a well-established technique in computer science known as *oversampling* (e.g., Chawla et al., 2002). Although this technique is widely used in computer science, to our knowledge it has not yet been applied by any personnel psychologists or management scholars to address issues of adverse impact. Oversampling, in its most basic form, involves resampling from the minority class with replacement (i.e., bootstrapping). Originally, this was done in binary classification problems to resolve class imbalances and improve classification accuracy. For example, in the realm of occupational health, computer scientists have used oversampling to enhance the prediction of whether a construction accident is likely to result in death (Choi et al., 2020). Since the vast majority (~98%) of accidents in Choi et al.'s (2020) data did not result in death, the classifier can be 98% accurate by always predicting that accidents do not result in death—yet the classifier may be 0% accurate at predicting workplace deaths. By oversampling the minority class (i.e., workplace accidents that result in death) during training, the researchers were better able to identify the predictors of deadly construction accidents and improved their ability to predict accidental construction deaths.

The present case differs from the historical application of these approaches because, rather than a binary classification task where one of the two classes (e.g., pass/fail) is underrepresented, our study involves an imbalanced representation of the binary classes (i.e., screened in/screened out) across demographic groups. Specifically, the SRs differ across groups, causing AI ratios that favor the majority class (i.e., White applicants). Therefore, the 'minority class' in the present study can be considered both a) screened in minorities (corresponding to group mean differences) and b) minorities in general (corresponding to differential representation). We expect adjustments prior to ML model training are legal because Ricci v. Destafano (2009) established that adjustments to enhance fairness are legal during test design (but not during or after test administration; Barocas & Selbst, 2016).

Recently, Yan et al. (2020) investigated oversampling's effects on procedural fairness (i.e., bias) in automated interview personality and hireability assessments (Yan et al., 2020). Oversampling to equalize outcome means and group sample sizes reduced the amount of demographic information contained in (i.e., improved the statistical parity of) automatic personality and hireability assessments and decreased accuracy differences along both gender and racial lines. Overall, oversampling underrepresented groups by resampling with replacement was effective at reducing measurement bias (i.e., differential model accuracy across groups; Tay et al., 2022), but we still do not know how it might affect adverse impact. We use oversampling to adjust training data AI ratios and sample sizes. However, multiple oversampling techniques exist and may differ in their effects.

**Oversampling Techniques**

Although Yan et al. (2020) oversampled real observations (i.e., bootstrapping) to achieve data balance, much work in computer science has used the Synthetic Minority Over-sampling TEchnique (SMOTE; Chawla et al., 2002). Therefore, real or *synthetic* observations can be oversampled to achieve data balance. SMOTE involves generating synthetic observations similar (but not identical) to real ones, as opposed to oversampling real observations with replacement. To do so, SMOTE first uses the *p* predictors that a predictive model will use to identify the *k* nearest neighbors (via Euclidean distance) for each observation in the specified class. Then, SMOTE subtracts the feature vector (i.e., predictor values) of an observation from each of its nearest neighbors' feature vectors, then multiplies these differences by a random number between 0 and 1 and adds the resulting values to the observation's feature vector. This results in synthetic observations at random points on the lines between the features of two observations. Figure 3 provides a two-dimensional illustration of this process. The synthetic observations can then be randomly sampled, as specified by the user. Using synthetic observations that do not exactly duplicate real observations reduces overfitting, increasing the generalizability of the resulting model (Chawla et al., 2002).

SMOTE has been so successful that it is considered one of the most influential data preprocessing techniques in ML (Garcia et al., 2016). Indeed, SMOTE is considered a "de facto" method for working with imbalanced data and has helped both with binary and multilabel classification tasks (Fernandez et al., 2018). The original application of SMOTE was in bioinformatics (e.g., classifying calcifications in mammograms; Chawla et al., 2002), and SMOTE has since been applied to a wide variety of domains, including predicting the likelihood of student failure in college courses (Costa et al., 2017), predicting behavior from smartphone accelerometers (Chen & Shen, 2017), and detecting aggression and bullying on Twitter (Chatzakou et al., 2017).

For personnel data, some may feel there is something uncanny (Mori, 2012) about using synthetically generated observations. This may be because our field is unfamiliar with the approach, it may be difficult to explain the approach to a judge and jury, and we know little about how applicants will react to the use of such techniques. However, oversampling real observations also raises some concerns. For example, an ML model trained by oversampling real observations may be overfitted to them, thereby reducing generalizability (indeed, this is one of the main motivations for generating synthetic observations; Chawla et al., 2002). To address these concerns, we conduct an initial investigation into whether ML model adverse impact is differentially affected by oversampling real or synthetic observations.

## Method

Data are not available due to their proprietary nature. Data were analyzed using multiple packages in Python (Van Rossum & Drake, 2009) and primarily Scikit Learn (Pedregosa et al., 2011). The study design was not preregistered because the data are real-world selection data.

**Self-Report Survey Scores**

The self-report survey consisted of 16 proprietary multi-item, bipolar response scales that tap ways of working and ways of relating to people at work. The items are responded to on a six-point scale that has descriptive anchors at each end. The descriptive anchors were designed to be equally socially desirable, similar to the single item Big Five scale developed by Woods and Hampson (2005). The scales ranged in length from five to 10 items. According to the scale's technical manual, Cronbach's alpha ranged from .66 (follow through) to .88 (sociability). Additionally, the measures exhibited expected intercorrelations with relevant NEO Big Five measures. For example, accommodation correlated $r = .629$ with agreeableness, follow through correlated $r = .674$ with conscientiousness, and sociability correlated $r = .812$ with extraversion. The scales also exhibit criterion-related validity, as, for example, work intensity exhibited correlations ranging from $r = .11$ (in sales jobs) to $r = .21$ (in customer service jobs) with job performance ratings. Further, in practice the scales are generally combined into a single score based on mapping the scales to the knowledge, skills, abilities, and other characteristics required for the job (i.e., based on job analysis and O*NET data). The composite scores exhibited correlations with overall performance ratings ranging from a low of $r = .16$ (among 425 warehouse workers) and a high of $r = .25$ (among 360 laborers and freight, stock, and material movers, hand).

**Text Mined Interview Transcripts**

To vectorize transcripts, we used a variety of common approaches including Linguistic Inquiry and Word Count (LIWC; Pennebaker et al., 2015), counting one- and two-word phrases (i.e., $n$-grams when $n = 1$ & $2$), Latent Semantic Indexing (LSI; Deerwester et al., 1990), Latent Dirichlet Allocation (Blei et al., 2003), Doc2Vec (Le & Mikolov, 2014), and DistilBERT (Sanh et al., 2019). We used multiple techniques because we did not want our results to be bound to a specific technique. Additional details on each technique are provided below.

We also explored models that combined text mined interview transcripts with self-report scores. For those models, we combined one of the common approaches to vectorizing transcripts with the raw self-report scale scores. As reported in Tables 1 and S2-S7, those models tended not to be more accurate, on average, than models that only included the text mined interview transcripts. Additionally, although the raw values of the AI ratios observed in those models differed from the models that only used interview transcripts, the pattern of effects caused by oversampling was similar across the two sets of models.

*Text Preprocessing*

As part of the vectorization of natural language text data, some approaches involve extensive preprocessing of text. Where noted below, we implemented a preprocessing routine which involved removing all numbers, converting text to lowercase, handling negation, removing punctuation, removing extra white space, removing stop words (i.e., words so common that they are often uninformative), and stemming all words. These preprocessing steps follow recent recommendations (Hickman et al., 2022), with the exception of the use of stemming. We opted for stemming to further reduce the dimensionality of our data considering the relatively small sample size.

*Linguistic Inquiry and Word Count*

Linguistic Inquiry and Word Count includes a series of linguistic and psychological dictionaries comprised of words and phrases (Pennebaker et al., 2015). Dictionaries are scored as the proportion of total text that matches the words and phrases it contains. The LIWC models used 93 dictionaries, including both linguistic dictionaries (e.g., verbs, pronouns) and psychological ones (e.g., positive emotions; driven by achievement). We calculated LIWC scores at the interview level.

*n-grams*

We used the tm R package (Feinerer & Hornik, 2019) to count all *n*-grams where $n = 1$ or 2 (i.e., one- and two-word phrases) from the pre-processed text at the overall interview level. We then removed all *n*-grams that did not occur in at least 2% of the interviews.

*Latent Semantic Indexing (LSI)*

LSI (Deerwester et al., 1990) is an unsupervised mathematical technique based on Singular Value Decomposition (SVD) which brings out the 'latent' semantics of a corpus of documents. From the preprocessing transcripts, we counted the occurrence of each *n*-gram where $n = 1$, then we applied the term-frequency-inverse document frequency (TF-IDF) transformation to those counts. We then used the Gensim Python package (Rehurek & Sojka, 2010) to implement LSI and extract 50 latent topics from the text.

*Latent Dirichlet Allocation (LDA)*

LDA (Blei et al. 2003) is one of the most popular topic modeling algorithms, which is an unsupervised statistical methodology to learn "topics" from a large corpus of documents. This process is analogous to exploratory factor analysis and LSI, in that it seeks to inductively identify the latent topics in the text. To generate LDA vectors we first preprocessed the interview transcript as described above. Then, we counted all *n*-grams where *n* = 1 and applied the TF-IDF transformation. We then used the Gensim Python package (Rehurek & Sojka, 2010) to generate 50 topic vectors from the TF-IDF vectors.

*Doc2Vec*

The Doc2Vec algorithm (Le & Mikolov, 2014) transforms text to a fixed dimensional representation of its semantics by trying to predict words in a document. Using the Gensim Python package (Rehurek & Sojka, 2010), we trained Doc2Vec with a window size of 2 on the raw text (without preprocessing) from the training folds and then applied it to the test folds to create a 100-dimension representation of each interview's semantics.

*DistilBERT*

Using the transformers Python package (Wolf et al., 2020), we used DistilBERT to extract 768 parameters describing the semantics of speech (Sanh et al., 2019). DistilBERT utilizes BERT's pre-trained architecture and outputs the final layer of BERT's neural network without requiring additional finetuning. These parameters can be used in either deep learning or inputted to traditional ML algorithms, as done in the present study. Because DistilBERT is limited to a maximum of 512 tokens, we extracted DistilBERT parameters separately for each of the five interview questions and used the 768 parameters from each question as predictors.

**Predictive Algorithms**

We used multiple predictive algorithms, including logistic regression, ridge regression (Hoerl & Kennard, 1970), Gaussian Naive Bayes, linear discriminant analysis (Fisher, 1936), random forest (Breiman, 2001), AdaBoost (Freund & Schapire, 1997), gradient boosting (Friedman, 1999), *k* Nearest Neighbors (Fix & Hodges, 1989), decision trees (Breiman et al., 1984), quadratic discriminant analysis (Tharwat, 2016), and support vector machine (Boser et al., 1992) classifiers as implemented in Scikit Learn (Pedregosa et al., 2011). These models represent commonly used classification algorithms in machine learning.

We chose to use multiple algorithms to test the generalizability of the effects of training data adverse impact ratios across many algorithms, rather than risking that our results be bound to a given predictor-algorithm pairing. Readers interested in learning more about these classifiers should review the citations provided above and/or refer to research in psychology and management that has summarized these approaches (e.g., Hayes et al., 2015; Putka et al., 2018; Strobl et al., 2009) as well as machine learning textbooks which devote entire chapters to different types of predictive algorithms (e.g., Hastie et al., 2009).

For each classifier, we conducted hyperparameter tuning in each fold of the raw data (i.e., three times for each predictor-algorithm pair). We conducted a 5-fold cross-validation to identify optimal hyperparameters, using hyperparameters identified as important by Yang and Shami (2020). We used a halving random search as implemented in Scikit Learn due to its greater efficiency compared to grid and random search (Jamieson et al., 2016; Li et al., 2018). We then used those optimal hyperparameters for all remaining experiments on that fold with that predictor-algorithm pair. If we had conducted hyperparameter tuning for each predictor-algorithm pair in each experimental condition, this would represent a confound because more than one variable would change across experimental conditions. We report the hyperparameters that we tried during the inner 5-fold cross-validation below.

*Logistic Regression Hyperparameters*

We did not tune hyperparameters for logistic regression because it has none.

*Ridge Regression Hyperparameters*

We tried four values of alpha: 0.5, 1.0, 5, and 10.

*Gaussian Naïve Bayes Hyperparameters*

We tried three values of the smoothing parameter, alpha: $1^{-8}$, $1^{-9}$, and $1^{-10}$.

*Linear Discriminant Analysis Hyperparameters*

We tried three values of solver: singular value decomposition, least squares, and eigenvalues.

*Random Forest Hyperparameters*

We tried four values for the number of trees in the forest: 50, 100, 150, and 200. We tried three values for the criterion: GINI, entropy, and log loss. We tried five values for the maximum tree depth: no maximum, 5, 15, 30, and 60. We tried three values for the maximum number of predictors given to each tree: no maximum, $\log(k)$, and $k^{\wedge}.5$ (where $k$ is the total number of predictors given to the model).

*AdaBoost Hyperparameters*

We tried four values for the number of estimators: 25, 50, 100, and 150. We tried three values for the learning rate: 0.75, 1.0, and 1.5.

*Gradient Boosting Hyperparameters*

We tried three values for the number of estimators: 50, 100, and 200. We tried three values for the learning rate: .05, .10, and .20.

*k Nearest Neighbors Hyperparameters*

We tried four values for the number of neighbors: 3, 5, 10, and 15. We tried to metrics for estimating distance: Minkowski and Euclidean.

*Decision Trees Hyperparameters*

We tried four values for the maximum tree depth: no maximum, 3, 8, and 15. We tried three values for the criterion: GINI, entropy, and log loss.

*Quadratic Discriminant Analysis Hyperparameters*

We tried three values for the regularization parameter: 0, 1, and 5.

*Support Vector Machine Hyperparameters*

We tried four types of kernels: linear, polynomial, Gaussian, and sigmoid. We tried three values for the degree parameter: 1, 3, and 5.

**Online Supplement References**

ONLINE SUPPLEMENT                                                                 13application, and characteristics of classification and regression trees, bagging, and random forests. *Psychological Methods*, *14*(4), 323-348.

Tharwat, A. (2016). Linear vs. quadratic discriminant analysis classifier: A tutorial. *International Journal of Applied Pattern Recognition*, *3*(2), 145-180.

Van Rossum, G., & Drake, F. L. (2009). *Python 2.6 Reference Manual*.

Wolf, T., Chaumond, J., Debut, L., Sanh, V., Delangue, C., Moi, A., ... & Rush, A. M. (2020, October). Transformers: State-of-the-art natural language processing. In *Proceedings of the 2020 Conference on Empirical Methods in Natural Language Processing: System Demonstrations* (pp. 38-45).

Woods, S. A., & Hampson, S. E. (2005). Measuring the Big Five with single items using a bipolar response scale. *European Journal of Personality: Published for the European Association of Personality Psychology*, *19*(5), 373-390.

Yan, S., Huang, D., & Soleymani, M. (2020, October). Mitigating biases in multimodal personality assessment. In *Proceedings of the 2020 International Conference on Multimodal Interaction* (pp. 361-369).

Yang, L., & Shami, A. (2020). On hyperparameter optimization of machine learning algorithms: Theory and practice. *Neurocomputing*, *415*, 295-316.

## Online Supplement Tables

**Table S1**

*Correlations Between Experimental Conditions and ML Model Adverse Impact Ratios (Overall SR = .50)*

|  | Self-reports | | | Interview Transcripts | | | Combined Predictors | | |
| --- | --- | --- | --- | --- | --- | --- | --- | --- | --- |
|  | NW | B | H | NW | B | H | NW | B | H |
| Training Data AI Ratio | .42 | .45 | .29 | .18 | .22 | .11 | .24 | .29 | .16 |
| SRs vs. SRs & *N*s | .05 | .06 | .03 | .02 | .02 | .03 | .03 | .02 | .03 |
| Real vs. Synthetic | .03 | .05 | .00 | -.02 | -.00 | -.02 | -.03 | -.02 | -.01 |

*Note*: NW = non-White/White AI ratios; B = Black/White AI ratios; H = Hispanic/White AI ratios. SRs vs. SRs & *N*s coded such that SRs = 0, SRs & *N*s = 1. Real vs. Synthetic coded such that Real = 0, Synthetic = 1. The values reported are the correlation between ML model AI ratios in test data, the manipulated training data AI ratios, and dummy variables for the remaining experimental conditions. For self-reports, *N* = 1,320 models; for interview transcripts, *N* = 3,960 models; for both predictor sets, *N* = 3,960 models.

**Table S2**

*Average ML Model Accuracy and Adverse Impact Ratios in Experimental Conditions: Self-Reports (Overall SR = .50)*

|                |           | Accuracy |       |       |          | AI Ratio |      |      |
|----------------|-----------|----------|-------|-------|----------|----------|------|------|
| Train AI Ratio | Condition | Overall  | White | Black | Hispanic | NW/W     | B/W  | H/W  |
|                | Raw       | .568     | .584  | .558  | .569     | .736     | .692 | .753 |
| .6             | All       | .564     | .581  | .552  | .563     | .724     | .689 | .737 |
|                | SRs       | .565     | .583  | .551  | .563     | .720     | .682 | .735 |
|                | SRs & *N*s | .563    | .579  | .552  | .563     | .728     | .696 | .740 |
|                | Real      | .564     | .581  | .551  | .563     | .720     | .683 | .734 |
|                | Synthetic | .564     | .582  | .552  | .563     | .727     | .695 | .741 |
| .8             | All       | .564     | .577  | .554  | .562     | .743     | .711 | .754 |
|                | SRs       | .565     | .579  | .553  | .565     | .733     | .695 | .746 |
|                | SRs & *N*s | .563    | .575  | .555  | .560     | .754     | .727 | .761 |
|                | Real      | .563     | .577  | .554  | .562     | .741     | .709 | .754 |
|                | Synthetic | .564     | .578  | .554  | .562     | .745     | .713 | .754 |
| 1.0            | All       | .560     | .570  | .551  | .560     | .768     | .745 | .771 |
|                | SRs       | .561     | .571  | .551  | .558     | .765     | .739 | .772 |
|                | SRs & *N*s | .560    | .569  | .551  | .561     | .771     | .751 | .770 |
|                | Real      | .560     | .570  | .550  | .557     | .766     | .739 | .771 |
|                | Synthetic | .561     | .570  | .552  | .562     | .770     | .751 | .770 |
| 1.2            | All       | .554     | .560  | .547  | .552     | .807     | .789 | .808 |
|                | SRs       | .554     | .559  | .547  | .552     | .808     | .791 | .810 |
|                | SRs & *N*s | .555    | .560  | .547  | .552     | .805     | .787 | .806 |
|                | Real      | .553     | .558  | .547  | .550     | .798     | .779 | .803 |
|                | Synthetic | .556     | .561  | .548  | .554     | .815     | .799 | .813 |
| 1.4            | All       | .544     | .545  | .541  | .540     | .854     | .844 | .849 |
|                | SRs       | .544     | .545  | .541  | .544     | .845     | .836 | .838 |
|                | SRs & *N*s | .544    | .544  | .542  | .536     | .863     | .851 | .861 |
|                | Real      | .543     | .545  | .538  | .538     | .855     | .839 | .857 |
|                | Synthetic | .546     | .544  | .544  | .542     | .853     | .849 | .841 |

**Table S3**

*Average ML Model Accuracy and Adverse Impact Ratios in Experimental Conditions: Interview Transcripts (Overall SR = .50)*

| Train AI Ratio | Condition | Accuracy | | | | AI Ratio | | |
|---|---|---|---|---|---|---|---|---|
| | | Overall | White | Black | Hispanic | NW/W | B/W | H/W |
| Raw | | .625 | .622 | .623 | .636 | .844 | .869 | .795 |
| .6 | All | .630 | .628 | .629 | .643 | .835 | .858 | .787 |
| | SRs | .630 | .628 | .629 | .643 | .833 | .853 | .784 |
| | SRs & *N*s | .630 | .629 | .630 | .642 | .837 | .862 | .789 |
| | Real | .631 | .629 | .630 | .643 | .837 | .860 | .788 |
| | Synthetic | .630 | .627 | .629 | .642 | .833 | .856 | .786 |
| .8 | All | .630 | .628 | .630 | .641 | .845 | .872 | .793 |
| | SRs | .630 | .628 | .630 | .641 | .846 | .874 | .792 |
| | SRs & *N*s | .630 | .628 | .630 | .640 | .844 | .870 | .793 |
| | Real | .630 | .628 | .630 | .640 | .846 | .872 | .794 |
| | Synthetic | .630 | .628 | .629 | .641 | .844 | .872 | .791 |
| 1.0 | All | .627 | .625 | .628 | .636 | .853 | .884 | .795 |
| | SRs | .627 | .625 | .628 | .636 | .853 | .887 | .793 |
| | SRs & *N*s | .627 | .625 | .629 | .637 | .853 | .881 | .797 |
| | Real | .629 | .625 | .629 | .639 | .856 | .887 | .798 |
| | Synthetic | .626 | .624 | .628 | .634 | .849 | .881 | .792 |
| 1.2 | All | .620 | .617 | .620 | .629 | .867 | .902 | .809 |
| | SRs | .620 | .617 | .620 | .629 | .864 | .900 | .804 |
| | SRs & *N*s | .620 | .617 | .621 | .628 | .870 | .904 | .814 |
| | Real | .624 | .620 | .625 | .633 | .869 | .901 | .814 |
| | Synthetic | .616 | .615 | .615 | .624 | .865 | .903 | .804 |
| 1.4 | All | .611 | .608 | .612 | .620 | .891 | .928 | .831 |
| | SRs | .612 | .609 | .613 | .621 | .882 | .919 | .820 |
| | SRs & *N*s | .611 | .607 | .612 | .620 | .899 | .937 | .841 |
| | Real | .618 | .612 | .619 | .627 | .891 | .926 | .832 |
| | Synthetic | .605 | .603 | .606 | .613 | .890 | .930 | .829 |

**Table S4**

*Average ML Model Accuracy and Adverse Impact Ratios in Experimental Conditions: Combined Predictor Set (Overall SR = .50)*

|                |            | Accuracy |       |       |          | AI Ratio |      |      |
|----------------|------------|----------|-------|-------|----------|----------|------|------|
| Train AI Ratio | Conditions | Overall  | White | Black | Hispanic | NW/W     | B/W  | H/W  |
| Raw            |            | .633     | .636  | .628  | .639     | .782     | .798 | .739 |
| .6             | All        | .631     | .635  | .624  | .642     | .768     | .785 | .726 |
|                | SRs        | .632     | .635  | .625  | .644     | .767     | .785 | .723 |
|                | SRs & *N*s | .630     | .635  | .623  | .640     | .769     | .785 | .728 |
|                | Real       | .631     | .634  | .624  | .643     | .772     | .788 | .727 |
|                | Synthetic  | .631     | .635  | .623  | .642     | .765     | .782 | .724 |
| .8             | All        | .631     | .633  | .627  | .640     | .781     | .801 | .735 |
|                | SRs        | .631     | .633  | .627  | .641     | .781     | .800 | .735 |
|                | SRs & *N*s | .631     | .634  | .627  | .640     | .782     | .802 | .736 |
|                | Real       | .631     | .633  | .627  | .642     | .784     | .803 | .738 |
|                | Synthetic  | .631     | .633  | .627  | .639     | .779     | .799 | .733 |
| 1.0            | All        | .628     | .629  | .624  | .637     | .799     | .825 | .747 |
|                | SRs        | .628     | .629  | .624  | .639     | .798     | .827 | .743 |
|                | SRs & *N*s | .628     | .630  | .624  | .636     | .801     | .823 | .751 |
|                | Real       | .629     | .630  | .625  | .639     | .802     | .828 | .749 |
|                | Synthetic  | .627     | .629  | .623  | .636     | .796     | .821 | .746 |
| 1.2            | All        | .623     | .623  | .621  | .631     | .818     | .848 | .762 |
|                | SRs        | .623     | .623  | .620  | .631     | .815     | .846 | .759 |
|                | SRs & *N*s | .623     | .623  | .621  | .630     | .820     | .850 | .765 |
|                | Real       | .626     | .626  | .624  | .633     | .823     | .853 | .766 |
|                | Synthetic  | .620     | .620  | .617  | .628     | .812     | .843 | .757 |
| 1.4            | All        | .613     | .613  | .612  | .619     | .850     | .844 | .793 |
|                | SRs        | .614     | .614  | .611  | .619     | .838     | .873 | .781 |
|                | SRs & *N*s | .613     | .612  | .612  | .620     | .862     | .895 | .806 |
|                | Real       | .619     | .618  | .618  | .625     | .853     | .886 | .795 |
|                | Synthetic  | .608     | .608  | .605  | .614     | .847     | .883 | .792 |

**Table S5**

*Average ML Model Accuracy and Adverse Impact Ratios in Experimental Conditions: Self-Reports (Overall SR = .10)*

| Train AI Ratio | Conditions | Accuracy | | | | AI Ratio | | |
|---|---|---|---|---|---|---|---|---|
| | | Overall | White | Black | Hispanic | NW/W | B/W | H/W |
| Raw | | .523 | .446 | .541 | .617 | .593 | .561 | .585 |
| .6 | All | .520 | .440 | .541 | .615 | .593 | .555 | .606 |
| | SRs | .521 | .441 | .540 | .616 | .595 | .558 | .598 |
| | SRs & *N*s | .520 | .440 | .541 | .615 | .590 | .552 | .614 |
| | Real | .521 | .442 | .541 | .615 | .596 | .569 | .601 |
| | Synthetic | .520 | .439 | .540 | .616 | .589 | .541 | .610 |
| .8 | All | .521 | .442 | .543 | .614 | .586 | .563 | .594 |
| | SRs | .522 | .444 | .544 | .615 | .583 | .557 | .591 |
| | SRs & *N*s | .520 | .441 | .542 | .614 | .589 | .569 | .597 |
| | Real | .521 | .442 | .542 | .615 | .587 | .565 | .586 |
| | Synthetic | .521 | .442 | .543 | .614 | .586 | .562 | .602 |
| 1.0 | All | .520 | .441 | .545 | .614 | .630 | .620 | .615 |
| | SRs | .522 | .443 | .546 | .615 | .627 | .616 | .621 |
| | SRs & *N*s | .519 | .440 | .544 | .613 | .632 | .624 | .610 |
| | Real | .521 | .442 | .545 | .614 | .634 | .615 | .612 |
| | Synthetic | .520 | .441 | .545 | .614 | .625 | .625 | .619 |
| 1.2 | All | .517 | .436 | .545 | .610 | .706 | .717 | .680 |
| | SRs | .517 | .436 | .545 | .610 | .714 | .727 | .687 |
| | SRs & *N*s | .517 | .436 | .545 | .610 | .698 | .707 | .673 |
| | Real | .517 | .435 | .544 | .609 | .687 | .695 | .655 |
| | Synthetic | .518 | .436 | .547 | .611 | .725 | .738 | .706 |
| 1.4 | All | .514 | .431 | .542 | .606 | .784 | .808 | .756 |
| | SRs | .513 | .430 | .542 | .606 | .757 | .784 | .730 |
| | SRs & *N*s | .515 | .432 | .542 | .606 | .810 | .831 | .782 |
| | Real | .513 | .432 | .540 | .605 | .772 | .790 | .752 |
| | Synthetic | .514 | .431 | .544 | .607 | .795 | .825 | .759 |

**Table S6**

*Average ML Model Accuracy and Adverse Impact Ratios in Experimental Conditions: Interview Transcripts (Overall SR = .10)*

| Train AI Ratio | Conditions | Accuracy | | | | AI Ratio | | |
|---|---|---|---|---|---|---|---|---|
| | | Overall | White | Black | Hispanic | NW/W | B/W | H/W |
| Raw | | .540 | .462 | .565 | .633 | .729 | .753 | .690 |
| .6 | All | .543 | .466 | .566 | .637 | .707 | .741 | .651 |
| | SRs | .543 | .466 | .566 | .637 | .702 | .738 | .650 |
| | SRs & *N*s | .543 | .466 | .566 | .637 | .712 | .744 | .653 |
| | Real | .543 | .466 | .566 | .637 | .698 | .737 | .639 |
| | Synthetic | .543 | .466 | .566 | .637 | .716 | .745 | .664 |
| .8 | All | .542 | .465 | .566 | .635 | .726 | .771 | .665 |
| | SRs | .543 | .465 | .567 | .636 | .722 | .774 | .655 |
| | SRs & *N*s | .542 | .464 | .566 | .634 | .730 | .767 | .676 |
| | Real | .542 | .465 | .566 | .635 | .727 | .771 | .668 |
| | Synthetic | .542 | .464 | .567 | .635 | .726 | .770 | .663 |
| 1.0 | All | .541 | .461 | .567 | .633 | .756 | .815 | .685 |
| | SRs | .541 | .462 | .567 | .633 | .755 | .818 | .682 |
| | SRs & *N*s | .541 | .461 | .567 | .633 | .758 | .813 | .689 |
| | Real | .541 | .462 | .568 | .634 | .769 | .830 | .705 |
| | Synthetic | .540 | .461 | .566 | .633 | .744 | .801 | .666 |
| 1.2 | All | .538 | .458 | .565 | .632 | .786 | .850 | .706 |
| | SRs | .538 | .457 | .565 | .632 | .781 | .846 | .705 |
| | SRs & *N*s | .538 | .458 | .565 | .633 | .792 | .854 | .706 |
| | Real | .539 | .459 | .566 | .633 | .798 | .872 | .708 |
| | Synthetic | .537 | .456 | .563 | .632 | .775 | .828 | .703 |
| 1.4 | All | .534 | .452 | .563 | .629 | .825 | .897 | .739 |
| | SRs | .534 | .452 | .562 | .628 | .827 | .897 | .743 |
| | SRs & *N*s | .535 | .453 | .564 | .631 | .824 | .898 | .735 |
| | Real | .537 | .455 | .566 | .630 | .831 | .909 | .742 |
| | Synthetic | .532 | .449 | .560 | .628 | .819 | .886 | .736 |

**Table S7**

*Average ML Model Accuracy and Adverse Impact Ratios in Experimental Conditions: Combined Predictors (Overall SR = .10)*

| Train AI Ratio | Conditions | Accuracy | | | | AI Ratio | | |
|---|---|---|---|---|---|---|---|---|
| | | Overall | White | Black | Hispanic | NW/W | B/W | H/W |
| Raw | | | | | | | | |
| .6 | All | .545 | .470 | .566 | .637 | .608 | .624 | .588 |
| | SRs | .546 | .470 | .566 | .638 | .606 | .623 | .584 |
| | SRs & *N*s | .545 | .469 | .566 | .636 | .610 | .625 | .591 |
| | Real | .545 | .470 | .566 | .637 | .607 | .622 | .587 |
| | Synthetic | .545 | .470 | .566 | .638 | .609 | .627 | .589 |
| .8 | All | .545 | .469 | .567 | .635 | .621 | .649 | .588 |
| | SRs | .545 | .470 | .568 | .635 | .618 | .649 | .579 |
| | SRs & *N*s | .544 | .468 | .567 | .635 | .623 | .648 | .598 |
| | Real | .545 | .469 | .567 | .635 | .626 | .654 | .590 |
| | Synthetic | .545 | .469 | .567 | .635 | .616 | .644 | .587 |
| 1.0 | All | .543 | .466 | .568 | .633 | .654 | .694 | .607 |
| | SRs | .544 | .466 | .568 | .634 | .648 | .683 | .604 |
| | SRs & *N*s | .543 | .466 | .568 | .633 | .660 | .704 | .611 |
| | Real | .544 | .467 | .568 | .634 | .659 | .703 | .608 |
| | Synthetic | .543 | .465 | .568 | .633 | .649 | .684 | .606 |
| 1.2 | All | .541 | .461 | .568 | .631 | .691 | .740 | .640 |
| | SRs | .541 | .460 | .568 | .630 | .687 | .735 | .638 |
| | SRs & *N*s | .541 | .461 | .568 | .632 | .695 | .744 | .641 |
| | Real | .542 | .463 | .570 | .631 | .698 | .758 | .634 |
| | Synthetic | .539 | .459 | .567 | .631 | .685 | .722 | .646 |
| 1.4 | All | .537 | .456 | .567 | .629 | .752 | .819 | .677 |
| | SRs | .537 | .454 | .566 | .628 | .738 | .799 | .672 |
| | SRs & *N*s | .538 | .457 | .567 | .630 | .767 | .838 | .682 |
| | Real | .539 | .457 | .568 | .630 | .759 | .831 | .675 |
| | Synthetic | .539 | .457 | .568 | .630 | .759 | .831 | .675 |

**Table S8**

*Average Accuracy of Each Algorithm Used Across All Predictor Sets in the Raw Data (Overall SR = .50)*

| Algorithm | Accuracy |
|---|---|
| Decision Trees | 0.58 |
| K-nearest Neighbors | 0.57 |
| AdaBoost | 0.64 |
| Gaussian Naive Bayes | 0.64 |
| Quadratic Discriminant Analysis | 0.60 |
| Logistic Regression | 0.64 |
| Linear Discriminant Analysis | 0.62 |
| Support Vector Classifier | 0.58 |
| Ridge Regression | 0.64 |
| Gradient Boosting | 0.65 |
| Random Forest | 0.65 |

**Table S9**

*Average Accuracy of Each Predictor Set Used Across Algorithms in the Raw Data (Overall SR = .50)*

| Predictor Set | Accuracy |
|---|---|
| Self-Report Percentile Scores | 0.57 |
| Self-Report Raw Scores | 0.56 |
| Doc2Vec | 0.60 |
| Latent Dirichlet Allocation | 0.56 |
| Doc2Vec+Self-Report Raw Scores | 0.61 |
| Latent Dirichlet Allocation+Self-Report Raw Scores | 0.60 |
| DistilBERT+Self-Report Raw Scores | 0.65 |
| DistilBERT | 0.64 |
| Latent Semantic Indexing+Self-Report Raw Scores | 0.65 |
| Latent Semantic Indexing | 0.65 |
| *n*-grams+Self-Report Raw Scores | 0.64 |
| LIWC | 0.65 |
| *n*-grams | 0.64 |
| LIWC+Self-Report Raw Scores | 0.66 |

*Note.* LIWC = Linguistic Inquiry and Word Count.

**Table S10**

*Average Accuracy of Each Predictor Set-Algorithm Pairing in the Raw Data (Overall SR = .50; Sorted from Most to Least Accurate)*

| Predictor Set | Algorithm | Accuracy |
| --- | --- | --- |
| Latent Semantic Indexing+Self-Report Raw Scores | Linear Discriminant Analysis | 0.71 |
| Latent Semantic Indexing+Self-Report Raw Scores | Ridge Regression | 0.70 |
| Latent Semantic Indexing | Linear Discriminant Analysis | 0.70 |
| Latent Semantic Indexing | Ridge Regression | 0.70 |
| Latent Semantic Indexing | Gaussian Naive Bayes | 0.69 |
| n-grams | Random Forest | 0.69 |
| DistilBERT | Logistic Regression | 0.69 |
| DistilBERT+Self-Report Raw Scores | Gradient Boosting | 0.69 |
| Latent Semantic Indexing | Random Forest | 0.69 |
| LIWC | Random Forest | 0.69 |
| Latent Semantic Indexing+Self-Report Raw Scores | Random Forest | 0.69 |
| DistilBERT | Random Forest | 0.69 |
| LIWC | Linear Discriminant Analysis | 0.69 |
| Latent Semantic Indexing+Self-Report Raw Scores | Gradient Boosting | 0.69 |
| Latent Semantic Indexing | Logistic Regression | 0.69 |
| Latent Semantic Indexing | Quadratic Discriminant Analysis | 0.69 |
| n-grams+Self-Report Raw Scores | Random Forest | 0.69 |
| LIWC+Self-Report Raw Scores | Logistic Regression | 0.69 |
| LIWC | Ridge Regression | 0.69 |
| LIWC+Self-Report Raw Scores | Linear Discriminant Analysis | 0.69 |
| LIWC+Self-Report Raw Scores | Gradient Boosting | 0.69 |
| LIWC+Self-Report Raw Scores | Ridge Regression | 0.69 |
| n-grams | Gradient Boosting | 0.69 |
| LIWC+Self-Report Raw Scores | Random Forest | 0.69 |
| Latent Semantic Indexing+Self-Report Raw Scores | Gaussian Naive Bayes | 0.68 |
| Latent Semantic Indexing | Gradient Boosting | 0.68 |
| DistilBERT | Gradient Boosting | 0.68 |
| DistilBERT+Self-Report Raw Scores | Logistic Regression | 0.68 |
| LIWC | Gradient Boosting | 0.68 |
| DistilBERT+Self-Report Raw Scores | Ridge Regression | 0.68 |
| n-grams+Self-Report Raw Scores | AdaBoost | 0.67 |
| Latent Semantic Indexing+Self-Report Raw Scores | AdaBoost | 0.67 |
| n-grams | AdaBoost | 0.67 |
| n-grams | Gaussian Naive Bayes | 0.67 |
| n-grams+Self-Report Raw Scores | Gaussian Naive Bayes | 0.67 |
| LIWC | Gaussian Naive Bayes | 0.67 |
| DistilBERT+Self-Report Raw Scores | AdaBoost | 0.67 |
| DistilBERT+Self-Report Raw Scores | Random Forest | 0.67 |
| LIWC+Self-Report Raw Scores | Gaussian Naive Bayes | 0.67 |
| Latent Semantic Indexing+Self-Report Raw Scores | Logistic Regression | 0.67 |
| n-grams+Self-Report Raw Scores | Gradient Boosting | 0.67 |
| LIWC | Logistic Regression | 0.67 |
| n-grams+Self-Report Raw Scores | Support Vector Classifier | 0.67 |
| LIWC | Support Vector Classifier | 0.67 |
| LIWC | AdaBoost | 0.66 |
| LIWC+Self-Report Raw Scores | AdaBoost | 0.66 |
| Latent Dirichlet Allocation+Self-Report Raw Scores | Random Forest | 0.66 |
| Latent Dirichlet Allocation+Self-Report Raw Scores | Gradient Boosting | 0.66 |
| Latent Semantic Indexing | AdaBoost | 0.66 |
| DistilBERT | AdaBoost | 0.66 |
| DistilBERT | Ridge Regression | 0.66 |
| Latent Dirichlet Allocation+Self-Report Raw Scores | AdaBoost | 0.65 |
| n-grams | Support Vector Classifier | 0.65 |
| Doc2Vec | Gaussian Naive Bayes | 0.64 |
| Doc2Vec+Self-Report Raw Scores | Gaussian Naive Bayes | 0.64 |
| DistilBERT+Self-Report Raw Scores | Gaussian Naive Bayes | 0.64 |
| DistilBERT | Gaussian Naive Bayes | 0.64 |
| n-grams+Self-Report Raw Scores | Logistic Regression | 0.64 |
| Doc2Vec | Quadratic Discriminant Analysis | 0.64 |
| Latent Dirichlet Allocation | AdaBoost | 0.64 |
| Latent Dirichlet Allocation | Gradient Boosting | 0.64 |
| Latent Dirichlet Allocation | Random Forest | 0.64 |
| Doc2Vec+Self-Report Raw Scores | Logistic Regression | 0.64 |
| DistilBERT | Quadratic Discriminant Analysis | 0.64 |
| DistilBERT+Self-Report Raw Scores | Quadratic Discriminant Analysis | 0.64 |
| DistilBERT | Support Vector Classifier | 0.64 |
| DistilBERT+Self-Report Raw Scores | Support Vector Classifier | 0.64 |
| LIWC+Self-Report Raw Scores | Support Vector Classifier | 0.64 |
| Doc2Vec | Ridge Regression | 0.63 |
| n-grams | Logistic Regression | 0.63 |
| Doc2Vec | Logistic Regression | 0.63 |
| Doc2Vec+Self-Report Raw Scores | Gradient Boosting | 0.63 |
| Doc2Vec+Self-Report Raw Scores | Support Vector Classifier | 0.63 |
| n-grams+Self-Report Raw Scores | K-nearest Neighbors | 0.63 |
| Doc2Vec+Self-Report Raw Scores | Ridge Regression | 0.63 |
| n-grams | K-nearest Neighbors | 0.63 |
| n-grams | Quadratic Discriminant Analysis | 0.63 |
| n-grams | Decision Trees | 0.62 |
| Doc2Vec+Self-Report Raw Scores | Quadratic Discriminant Analysis | 0.62 |
| Latent Semantic Indexing | K-nearest Neighbors | 0.62 |
| n-grams+Self-Report Raw Scores | Ridge Regression | 0.61 |
| DistilBERT+Self-Report Raw Scores | Decision Trees | 0.61 |
| n-grams+Self-Report Raw Scores | Decision Trees | 0.61 |
| DistilBERT+Self-Report Raw Scores | Linear Discriminant Analysis | 0.61 |
| DistilBERT | Linear Discriminant Analysis | 0.61 |
| Doc2Vec+Self-Report Raw Scores | Linear Discriminant Analysis | 0.61 |
| DistilBERT+Self-Report Raw Scores | K-nearest Neighbors | 0.61 |
| LIWC | K-nearest Neighbors | 0.61 |
| LIWC+Self-Report Raw Scores | K-nearest Neighbors | 0.61 |
| LIWC+Self-Report Raw Scores | Decision Trees | 0.61 |
| Latent Semantic Indexing+Self-Report Raw Scores | Support Vector Classifier | 0.61 |
| n-grams | Ridge Regression | 0.60 |
| DistilBERT | Decision Trees | 0.60 |
| Self-Report Raw Scores | Linear Discriminant Analysis | 0.60 |
| Self-Report Raw Scores | Logistic Regression | 0.60 |
| Self-Report Raw Scores | Ridge Regression | 0.60 |
| Latent Dirichlet Allocation+Self-Report Raw Scores | Linear Discriminant Analysis | 0.60 |
| Latent Dirichlet Allocation+Self-Report Raw Scores | Ridge Regression | 0.60 |
| Latent Dirichlet Allocation+Self-Report Raw Scores | Logistic Regression | 0.60 |
| Doc2Vec+Self-Report Raw Scores | AdaBoost | 0.60 |
| DistilBERT | K-nearest Neighbors | 0.60 |
| LIWC | Decision Trees | 0.60 |
| Latent Semantic Indexing+Self-Report Raw Scores | Quadratic Discriminant Analysis | 0.60 |
| Latent Dirichlet Allocation+Self-Report Raw Scores | Decision Trees | 0.60 |
| n-grams+Self-Report Raw Scores | Linear Discriminant Analysis | 0.59 |
| Self-Report Percentile Scores | Support Vector Classifier | 0.59 |
| Doc2Vec | Linear Discriminant Analysis | 0.59 |
| Self-Report Percentile Scores | Linear Discriminant Analysis | 0.59 |
| Self-Report Percentile Scores | Ridge Regression | 0.59 |
| Self-Report Percentile Scores | Logistic Regression | 0.59 |
| Self-Report Raw Scores | Random Forest | 0.59 |
| Self-Report Percentile Scores | Random Forest | 0.59 |
| Self-Report Percentile Scores | Gradient Boosting | 0.59 |
| Self-Report Raw Scores | Gaussian Naive Bayes | 0.59 |
| Self-Report Raw Scores | AdaBoost | 0.59 |
| Doc2Vec+Self-Report Raw Scores | Random Forest | 0.59 |
| Self-Report Raw Scores | Gradient Boosting | 0.59 |
| Doc2Vec | Gradient Boosting | 0.59 |
| Doc2Vec | AdaBoost | 0.59 |
| n-grams | Linear Discriminant Analysis | 0.58 |
| Self-Report Percentile Scores | Quadratic Discriminant Analysis | 0.58 |
| Self-Report Raw Scores | Quadratic Discriminant Analysis | 0.58 |
| Latent Semantic Indexing | Decision Trees | 0.58 |
| Self-Report Percentile Scores | Gaussian Naive Bayes | 0.58 |
| Doc2Vec | Random Forest | 0.58 |
| LIWC+Self-Report Raw Scores | Quadratic Discriminant Analysis | 0.58 |
| LIWC | Quadratic Discriminant Analysis | 0.58 |
| Latent Dirichlet Allocation | Decision Trees | 0.58 |
| Doc2Vec | K-nearest Neighbors | 0.58 |
| Doc2Vec | Decision Trees | 0.57 |
| Doc2Vec+Self-Report Raw Scores | Decision Trees | 0.57 |
| Self-Report Percentile Scores | AdaBoost | 0.57 |
| Latent Semantic Indexing+Self-Report Raw Scores | Decision Trees | 0.57 |
| Latent Dirichlet Allocation+Self-Report Raw Scores | Gaussian Naive Bayes | 0.57 |
| Latent Dirichlet Allocation+Self-Report Raw Scores | Quadratic Discriminant Analysis | 0.57 |
| Latent Dirichlet Allocation | Ridge Regression | 0.56 |
| Latent Dirichlet Allocation | Logistic Regression | 0.56 |
| Latent Dirichlet Allocation | Linear Discriminant Analysis | 0.56 |
| Doc2Vec+Self-Report Raw Scores | K-nearest Neighbors | 0.56 |
| Latent Dirichlet Allocation | Quadratic Discriminant Analysis | 0.55 |
| n-grams+Self-Report Raw Scores | Quadratic Discriminant Analysis | 0.55 |
| Latent Dirichlet Allocation | Gaussian Naive Bayes | 0.54 |
| Latent Dirichlet Allocation+Self-Report Raw Scores | Support Vector Classifier | 0.53 |
| Doc2Vec | Support Vector Classifier | 0.53 |
| Self-Report Percentile Scores | Decision Trees | 0.52 |
| Self-Report Percentile Scores | K-nearest Neighbors | 0.52 |
| Latent Semantic Indexing+Self-Report Raw Scores | K-nearest Neighbors | 0.52 |
| Self-Report Raw Scores | K-nearest Neighbors | 0.52 |
| Self-Report Raw Scores | Decision Trees | 0.51 |
| Latent Dirichlet Allocation+Self-Report Raw Scores | K-nearest Neighbors | 0.51 |
| Latent Dirichlet Allocation | K-nearest Neighbors | 0.50 |
| Self-Report Raw Scores | Support Vector Classifier | 0.45 |
| Latent Semantic Indexing | Support Vector Classifier | 0.45 |
| Latent Dirichlet Allocation | Support Vector Classifier | 0.39 |

*Note.* LIWC = Linguistic Inquiry and Word Count.

**Online Supplement Figures**

**Figure S1**

*Machine Learning Model Development Process*

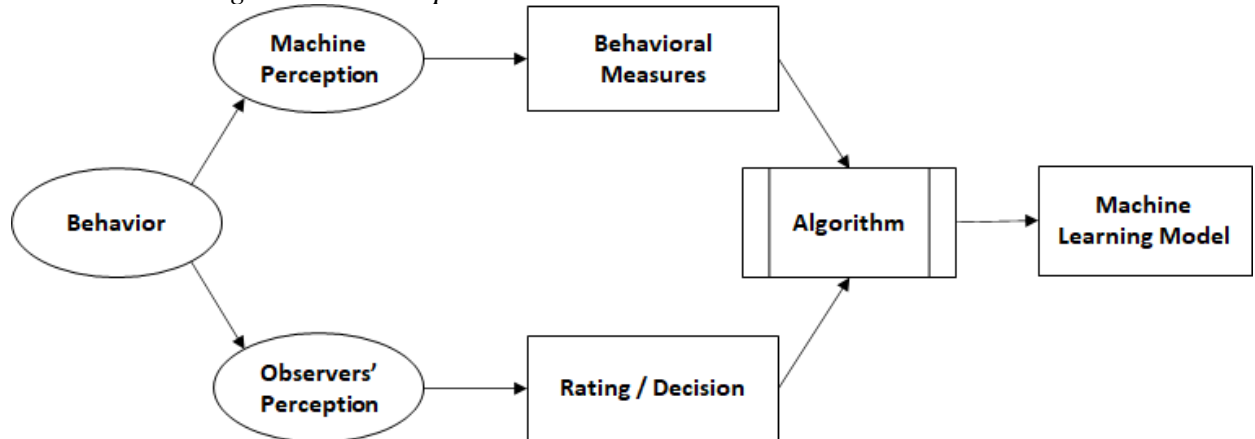

*Note*: In this case, behavior is either interview performance or responses to a self-report instrument or a test. Machine perception, therefore, can either be converting unstructured behavior into structured data or scoring self-report scales (e.g., with summing or ideal point). Observers' perception may involve only observing this behavior or could be a more holistic process of judgment and decision-making based on multiple inputs. The algorithm is then fed the behavioral measures to predict the human rating/decision, resulting in an ML model that can be used to automatically predict the rating/decision on future cases.

RESPONSE LETTER

**Figure S2**

*Illustration of How SMOTE Creates Synthetic Observations*

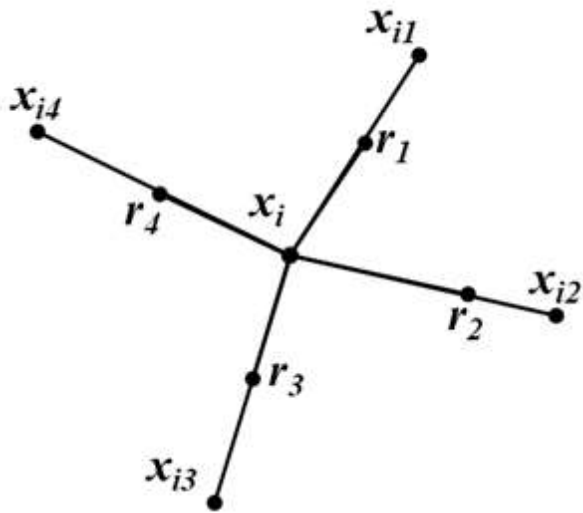

*Note*: Adapted from Fernandez et al. (2018). $x_i$ = original observation, $x_{i1\text{-}4}$ = the four nearest neighbors, and $r_{1\text{-}4}$ = synthetic observations generated from the original observation and its neighbors.